# Interpretable AI-driven Guidelines for Type 2 Diabetes Treatment from Observational Data

Dewang K. Agarwal[1], Dimitris J. Bertsimas[1]

[1] Sloan School of Management and Operations Research Center, E62-560, Massachusetts Institute of Technology, Boston, MA, USA

Keywords: Type 2 Diabetes Treatment, Treatment Decision Making, Treatment Algorithms, Observational Study, Artificial Intelligence, Data Analysis, Machine Learning

**Abstract**

**Objective:** Create precise, structured, data-backed guidelines for type 2 diabetes treatment progression, suitable for clinical adoption.

**Research Design and Methods:** Our training cohort was composed of patient (with type 2 diabetes) visits from Boston Medical Center (BMC) from 1998 to 2014. We divide visits into 4 groups based on the patient's treatment regimen before the visit, and further divide them into subgroups based on the recommended treatment during the visit. Since each subgroup has observational data, which has confounding bias (sicker patients are prescribed more aggressive treatments), we used machine learning and optimization to remove some datapoints so that the remaining data resembles a randomized trial. On each subgroup, we train AI-backed tree-based models to prescribe treatment changes. Once we train these tree models, we manually combine the models for every group to create an end-to-end prescription pipeline for all patients in that group. In this process, we prioritize stepping up to a more aggressive treatment before considering less aggressive options. We tested this pipeline on unseen data from BMC, and an external dataset from Hartford healthcare (type 2 diabetes patient visits from January 2020 to May 2024).

**Results:** The median HbA1c reduction achieved by our pipelines is 0.26% more than what the doctors achieved on the unseen BMC patients. For the Hartford cohort, our pipelines were better by 0.13%.

**Conclusions:** This precise, interpretable, and efficient AI-backed approach to treatment progression in type 2 diabetes is predicted to outperform the current practice and can be deployed to improve patient outcomes.

About 10% of Americans suffer from diabetes. Out of these, 90%-95% are cases of type 2 diabetes (1) The impact of this disease is large, both from a health and an economic standpoint (2). Diet, lifestyle changes, exercise, weight loss are the foundations of all type 2 diabetes management regimens, but they must be supplemented by medications in almost all cases. Over the recent decades, there have been significant developments in type 2 diabetes medications with the introduction of multiple new medications, like DPP-4, GLP-1 agonists, SGLT-2 antagonists, thiazolidinediones, sulfonylureas etc. As a result, the American Diabetes Association and the European Association for the Study of Diabetes, came up with new guidelines for more individualized type 2 diabetes management (3,4).

Even with these established clinical guidelines, there is still significant discrepancy in the approach that doctors take to manage type 2 diabetes. This discrepancy is primarily due to variations in clinical experience, knowledge of medications, as well as patients' preferences. To streamline type 2 diabetes management, one of these studies suggested a cluster-based approach to determine the exact condition and treatment response of a patient (4). A related approach, nearest neighbors, has been studied in (5), where the proposed algorithm's recommendations were shown to reduce the HbA1c level in patients by 0.44% on average. While a nearest-neighbor-based approach is interpretable, it is not as interpretable and as structured as a policy for treatment prescription. Additionally, (5) does not validate their algorithm on an external dataset.

In this work, we use artificial intelligence (AI) and optimization to determine a prescriptive policy framework for treatment of type 2 diabetes. Specifically, we create smart end-to-end decision-making pipelines which prescribe regimens for patients at various stages of their treatment journeys. AI and machine learning techniques have been used in various aspects of type 2 diabetes ranging from the prediction of the likelihood of developing type 2 diabetes, to smart treatment strategies including smart diet monitoring and smart drug therapy, to even prediction of development of type 2 diabetes related co-morbidities (6). One of the papers mentioned in (6), uses sequential pattern mining to predict the next treatment for a patient (7). However, this work does not provide an interpretation of when should the next prescription be made for the patient. An interpretable algorithm for treatment was created by the American Association of Clinical Endocrinology (8). While this algorithm was developed through a consensus among clinicians, who are experts in prognosis of type 2 diabetes, yet there is no data to demonstrate the effectiveness of the algorithmic guidelines.

For our study, we use historical data from an electronic medical records (EMR) database same as (5). As pointed out by (5), there are various benefits to this EMR data like a large sample size and plentiful clinical evidence of the effect of multiple drugs. However, EMR data are observational (not randomized trial) data which suffer from confounding. Using recent advances in precision medicine (9), we address this shortcoming by deleting some of the observed data, so that the remaining data emulates data from a randomized trial.

It was found in (10), that even though patients and clinicians are willing to adopt AI in type 2 diabetes prescription, yet they would only do so if the AI is shown to be effective, and the model is interpretable enough for a human to understand its decision making. So, we primarily use tree-based models for their high interpretability. These models take in a patient's data which includes characteristics like age, sex, race, comorbidities, BMI history, and HbA1c history to determine a treatment policy to maximize reduction in HbA1c. We then combine these models into decision making pipelines which give an end-to-end recommendation to physicians. We demonstrate the effectiveness of these pipelines by comparing the predicted reduction in HbA1c had the patients followed the pipeline's recommendation, to the actual reduction in HbA1c observed when the patient followed their doctor's prescription. The effectiveness of these pipelines was validated using an external cohort, where we get similar results to our original cohort. Therefore, these pipelines are interpretable and effective.

In summary, our contributions include creating interpretable data-backed guidelines for treatment progression for type 2 diabetes. We create these guidelines using observational data while addressing the inherent shortcomings of observational data in the process. We also demonstrate the effectiveness of our proposed guidelines on external data.

## 2. Research Design and Methodology

### 2.1 Data

We provide a brief overview of the data below. For a detailed description of the data, we refer the readers to (5).

There are 22,387 (type 2 diabetes) patient visits in our dataset from Boston Medical Center from 1998 to 2014. We will refer to this dataset as the BMC cohort. We treat each visit as an independent event. The relevant features of each patient visit are the demographics of the patient (age, race, and sex), the last measured HbA1c and BMI levels, summary statistics of the historical HbA1c and BMI levels (25th percentile, median, mean, and 75th percentile for both HbA1c and BMI), and whether the patient has kidney disease (which is a contraindication to metformin). Note, that the $25^{th}$ percentile is the value such that 25% of the data falls below it, and similarly, $75^{th}$ percentile is the value such that 75% of the data falls below it. Beyond these features, we also have information on the patient's current treatment regimen, the doctor's prescribed regimen after the current visit, and the resulting HbA1c after the doctor's prescribed regimen.

We define the following inclusion criteria for the patient visits considered in this study:

1. The patient must be above 18 years of age at the time of the visit.
2. If the patient is not taking any type 2 diabetes-related medication currently, but either is pre-diabetic (last measured HbA1c of at least 5.7%) or has a 75th percentile of HbA1c of at least 7% (the patient had to take type 2 diabetes related medicine in the past).
3. If the patient is taking some type 2 diabetes related medication currently, but either is not treated (last measured HbA1c of at least 7%) or has a 75th percentile of HbA1c of at least 7% (the patient had to take type 2 diabetes related medicine in the past).
4. There was a decrease in HbA1c after following the doctor's prescription from the current visit.
5. The patient can be of any gender.

Every visit must follow all the above criteria. The fourth criterion is included to reduce the noise in the data. It filters out those times where the patient may not have taken the medicine correctly (or did not follow the non-medicine related measures suggested by their doctor). While this may result in exclusion of some cases when the patient's HbA1c increased even after adhering to the treatment regimen, yet there was no reliable method to distinguish the reason for an increase in HbA1c among the patients.

In the remainder of this work, we will use patient and patient visit interchangeably. This is because we have assumed that each visit is an independent event, and hence each new visit is like a new patient coming to a clinic.

At this stage, for every model we create separate training and testing datasets. For each model, we keep patients whose current regimen matches the subclass of patients studied by the model, and whose prescribed treatments match the targets that the model considers. Then this filtered data is randomly split into training and testing sets (80% and 20% respectively). To further de-noise the training data, we remove major outliers from the training set for each model independently by filtering out those rows where any (continuous) column's value may be over the 95th percentile of that column.

**2.2 Methods**

The methodology presented in this work has been greatly inspired by (9). We recommend referring to (9) for the detailed description of all the steps involved.

Observational data suffers from confounding bias which is seen as follows: patients with a more advanced type 2 diabetes will be prescribed a more aggressive treatment like insulin. For example, among patients who were on no medication, those recommended to step up to insulin monotherapy had an average last measured HbA1c of 9.81%, while those recommended to continue with no medication had an average last measured HbA1c of 8.71%. Clearly, the data are not like that in a randomized clinical trial.

To address this point, we develop a random forest model which predicts HbA1c based on patients that received a less aggressive treatment (for example, the patients on no medication; excluding any patient who was recommended insulin monotherapy). We then apply this model for the patients who received the more aggressive treatment. We group all the patients into 6 equally spaced buckets, covering the range of predicted HbA1c for the more aggressive treatment group, and the actual HbA1c for the less aggressive treatment group. In each bucket, we match a patient who received a less aggressive treatment with a "similar" patient who received the more aggressive treatment. That is, we match a patient who was prescribed no treatment with a patient with "similar" physiology who was prescribed insulin. This "similarity" is defined based on the patients' ages, HbA1c histories, and BMI histories. After matching we discard some of the least-similar patient pairings. Through this process, we obtain a dataset which resembles a clinical trial.

We then train a policy tree AI model on this processed dataset (11). The goal of this tree model is to maximize the reduction in HbA1c. While training these tree models, we weigh the more aggressive treatment class higher than the less aggressive treatment class. This addresses any residual confounding by making the AI model more conservative on the patients who received the more aggressive medication.

We use the above procedure to create 8 individual tree models which are grouped under 4 categories depending on the patient's current treatment regimen:

1. Patients on no medication
    a. Step up to insulin monotherapy tree
    b. Step up to some first line therapy tree
    c. Take metformin monotherapy or take some other hypoglycemic monotherapy tree
2. Patients on some other hypoglycemic monotherapy
    a. Step up to insulin plus one other hypoglycemic tree
    b. Step up to metformin plus one other hypoglycemic tree
3. Patients on metformin monotherapy
    a. Step up to metformin plus insulin tree
    b. Step up to metformin plus one other hypoglycemic tree
4. Patients on metformin plus one other hypoglycemic treatment
    a. Step up to metformin, insulin, and one other hypoglycemic tree

For each of the 4 groups, we manually connect the corresponding tree models into an end-to-end decision-making pipeline. The general algorithm followed in the manual connection process was to make decision to step up to a more aggressive treatment before making the decision to step up to a less aggressive treatment. For example, in the no medication pipeline (shown in Fig. 1), the decision to step up to insulin is made before the decision to step up to metformin monotherapy.

To evaluate our pipelines, we create ground truth models (GTMs) using the BMC cohort. For each combination of a patient's current treatment and recommended treatment during a visit, we train a GTM to predict the change in HbA1c. Then, we use these GTMs to predict the change in HbA1c for patients where our pipelines and the doctors disagreed.

We also evaluate our pipelines on an external dataset from Hartford Healthcare in Connecticut, USA. The Hartford cohort consisted of 17,482 patient (with type 2 diabetes) visits between January 2020 and May 2024. We obtained an institutional review board (IRB) approval to use this data for validating this study.

### 2.3 Data and Resource Availability

The datasets used in this study are not publicly available. Data may be anonymized and made available upon written request to the corresponding author. A comprehensive proposal for how the data will be used must be sent to the corresponding author for detailed assessment of the data request.

## 3. Results

Overall, on the BMC cohort there were 9,308 patient visits where the doctors and our pipelines disagreed. The median predicted reduction in HbA1c using our pipelines (1.06%) is 0.26% more than the median HbA1c reduction after following the doctor's prescription (0.80%). On the Hartford cohort, which is a completely external dataset, there were 11,409 patient visits where the doctors and our pipelines disagreed. On the Hartford cohort, the median predicted reduction in HbA1c using our pipelines (0.63%) is 0.13% more than the median HbA1c reduction after following the doctor's prescription (0.50%). We study each pipeline in detail in their respective subsections.

### 3.1 No medication pipeline

This pipeline is made of three constituent binary policy trees. These trees are:

1. *Step up to insulin monotherapy tree*: This tree divided the patients into 2 subgroups: those with a last measured HbA1c of at least 8.05% were advised to step up to insulin monotherapy, while those with last measured HbA1c of less than 8.05% were not. Medically, this policy suggests prescribing insulin monotherapy to patients who are on no medication and whose last measured HbA1c is high (at least 8.05%).

2. *Step up to some first line therapy tree*: This tree divided the patients into 2 subgroups: those with a 25th percentile of HbA1c of at least 6.013% were advised to step up to some first line therapy, while those with a 25th percentile of HbA1c of less than 6.013% were not. Medically, this policy suggests prescribing some first line therapy to patients who are not taking any medication and who have had an HbA1c of more than 6.013% in at least 3 out of 4 measurements.

3. *Take metformin monotherapy or take some other hypoglycemic monotherapy tree*: This tree divided the patients into 4 subgroups: one subgroup was suggested to step up to metformin monotherapy, while the others were recommended some other hypoglycemic. Patients with kidney-related contraindication to metformin, or with a last measured BMI of at least 37.02, or with a last measured HbA1c of less than 6.85%, were prescribed some other hypoglycemic. Those with no contraindication, a last measured BMI of less than 37.02, and a last measured HbA1c value of at least 6.85%, were recommended metformin monotherapy. Medically, one can interpret this policy as: prescribe one other hypoglycemic to patients who are currently on no medication, and who either have kidney disease, or were moderately or severely obese, or were diabetic but did not have a high HbA1c (at most 6.85%). All other patients are suggested metformin monotherapy.

The combined pipeline is illustrated in Fig. 1. When testing on the BMC cohort, the recommendation of this pipeline is predicted to lead to a median decrease in HbA1c of 1.28%. This is 0.38% higher than the median reduction in HbA1c of 0.90% achieved by the doctors. On the Hartford cohort, the recommendation of this pipeline is predicted to lead to a median decrease in HbA1c of 0.61%, which is 0.11% more than the median reduction in HbA1c of 0.50% achieved by the doctors.

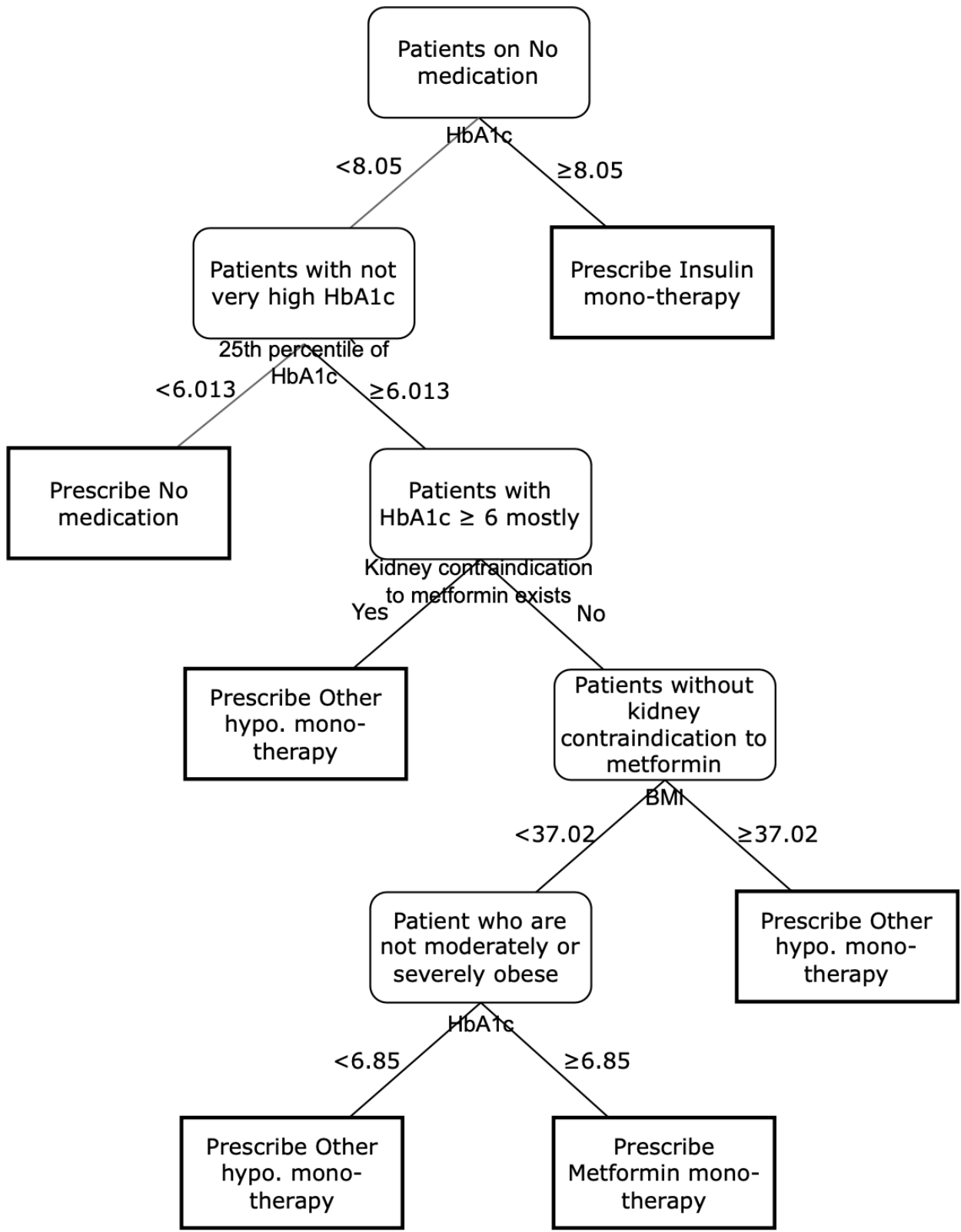

*Figure 1. Pipeline for patients on no medication. Hypo. refers to hypoglycemic.*

## 3.2 Other hypoglycemic pipeline

This pipeline is made of two constituent binary policy trees. These trees are:

1. *Step up to insulin plus one other hypoglycemic tree*: This tree divided the patients into 2 subgroups: those with a last measured HbA1c of at least 9.05% were advised to step up to insulin plus one other hypoglycemic, while those with last measured HbA1c of less than 9.05% were not. Medically, this policy suggests prescribing insulin plus one other hypoglycemic to patients who are on some other hypoglycemic monotherapy and whose last measured HbA1c is very high (at least 9.05%).

2. *Step up to metformin plus one other hypoglycemic tree*: This tree divided patients into 3 subgroups: those with no contraindication to metformin due to kidney disease and a last measured HbA1c value of at least 7.85% were advised to step up to metformin plus one other hypoglycemic, while others were not. Patients not advised to step up either had a kidney-related contraindication to metformin, or an HbA1c of less than 7.85%. Medically, this policy suggests prescribing metformin with one other hypoglycemic to patients who are taking one oral hypoglycemic and had no contraindication to metformin, and whose last measured HbA1c is high (at least 7.85%).

The combined pipeline is illustrated in Fig. 2. When testing on the BMC cohort, the recommendation of this pipeline is predicted to lead to a median decrease in HbA1c of 1.52%. This is 0.42% higher than the median reduction in HbA1c of 1.10% achieved by the doctors. On the Hartford cohort, the recommendation of this pipeline is predicted to lead to a median decrease in HbA1c of 0.87%, which is 0.07% more than the median reduction in HbA1c of 0.80% achieved by the doctors.

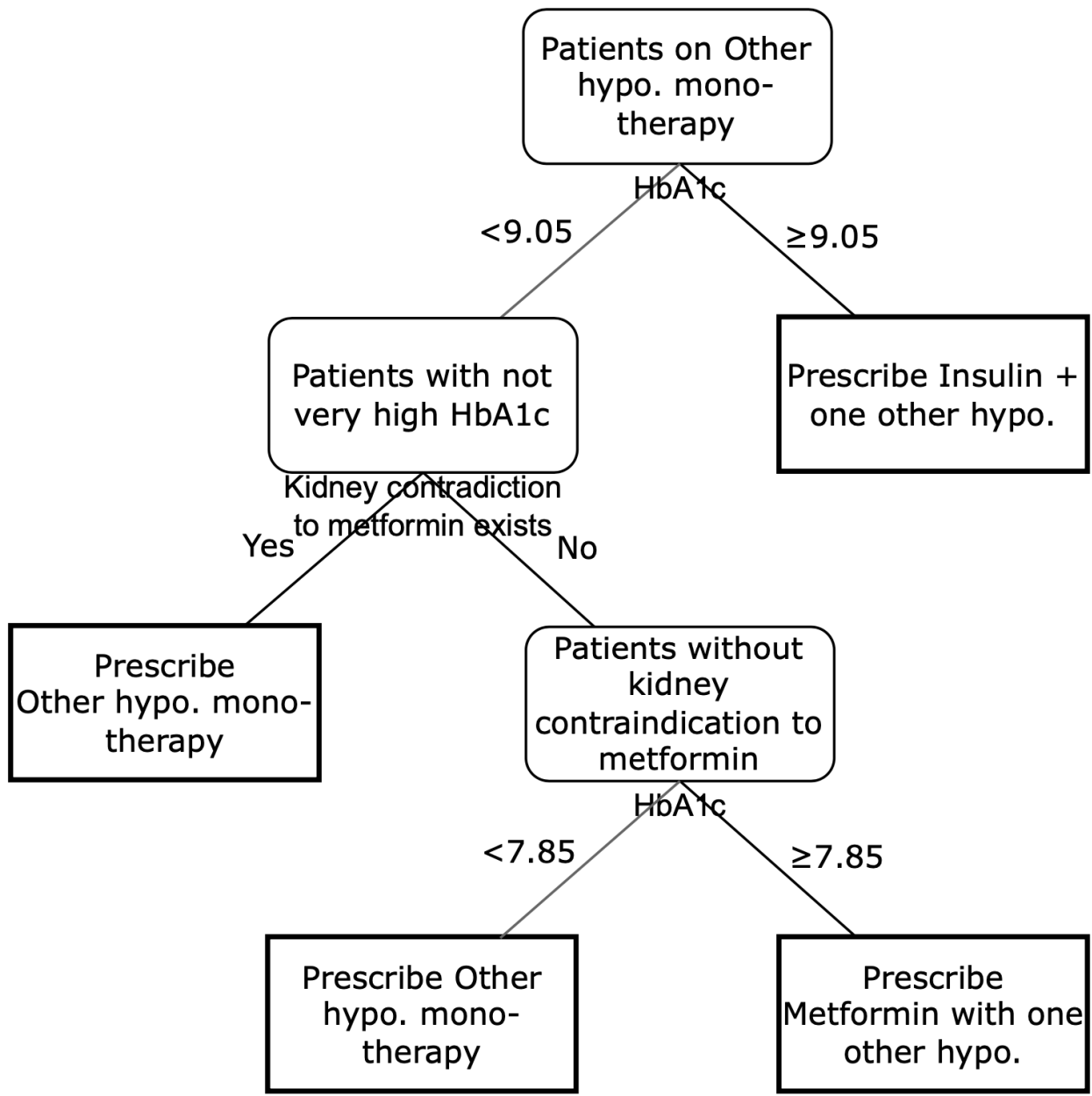

*Figure 2. Pipeline for patients on some other hypoglycemic monotherapy. Hypo. refers to hypoglycemic.*

### 3.3 Metformin pipeline

This pipeline is made of two constituent binary policy trees. These trees are:

1. *Step up to metformin plus insulin tree*: This tree divided the patients into 2 subgroups: those with a last measured HbA1c of at least 8.75% were advised to step up to metformin plus insulin, while those with last measured HbA1c of less than 8.75% were not. Medically, this policy suggests prescribing metformin plus insulin to patients who are taking metformin monotherapy and whose last measured HbA1c is very high (at least 8.75%).

2. *Step up to metformin plus one other hypoglycemic tree*: This tree divided the patients into 2 subgroups: those with a median BMI of at least 24.12 were advised to step up to metformin plus one other hypoglycemic, while those with median BMI of less than 24.12 were not. Medically, this policy suggests prescribing metformin with one other hypoglycemic to nearly overweight (having a BMI of more than 25) patients who are taking metformin monotherapy.

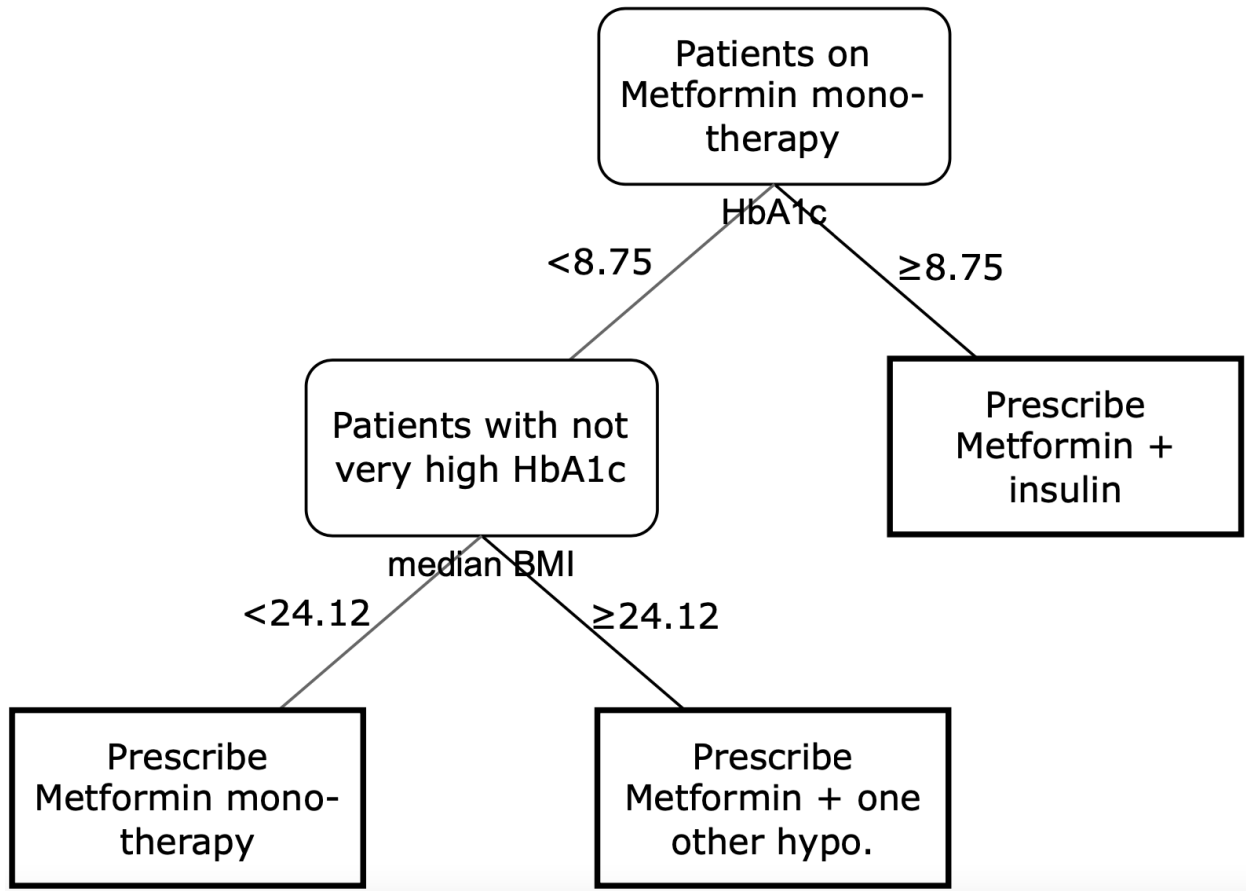

*Figure 3. Pipeline for patients on metformin monotherapy. Hypo. refers to hypoglycemic.*

The combined pipeline is illustrated in Fig. 3. When testing on the BMC cohort, the recommendation of this pipeline is predicted to lead to a median decrease in HbA1c of 0.82%. This is 0.12% higher than the median reduction in HbA1c of 0.70% achieved by the doctors. On the Hartford cohort, the recommendation of this pipeline is predicted to lead to a median decrease in HbA1c of 0.77%, which is 0.27% more than the median reduction in HbA1c of 0.50% achieved by the doctors.

### 3.4 Metformin plus one other hypoglycemic pipeline

This pipeline is made of two constituent binary policy trees. These trees are:

1. *Step up to metformin, insulin, and one other hypoglycemic tree*: This tree divided the patients into 2 subgroups: those with a median BMI of at least 27.35 were advised to step up to a metformin, insulin, and one other hypoglycemic, while those with median BMI of less than 27.35 were not. Medically, this policy suggests prescribing a metformin, insulin, and one other hypoglycemic to patients who are taking metformin monotherapy and are overweight or obese (having a BMI of more than 25 or 30 respectively).

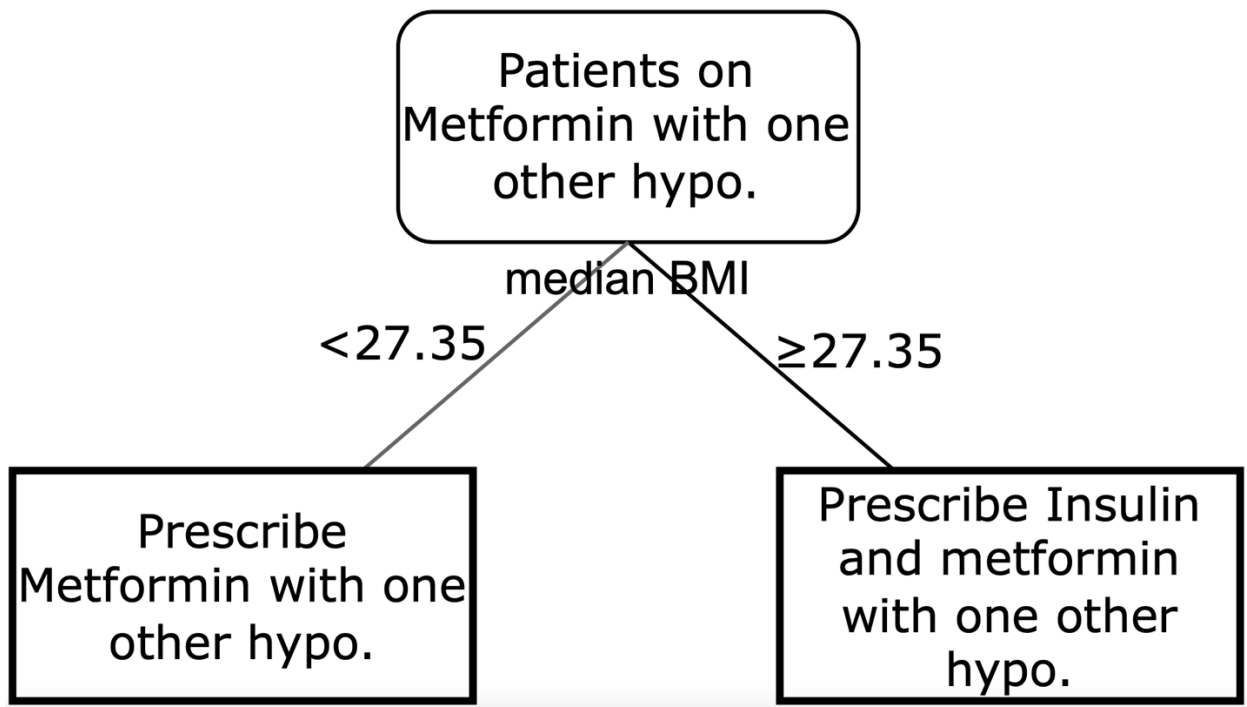

*Figure 4. Pipeline for patients on metformin with one other hypoglycemic treatment. Hypo. refers to hypoglycemic.*

The combined pipeline is illustrated in Fig. 4. When testing on the BMC cohort, the recommendation of this pipeline is predicted to lead to a median decrease in HbA1c of 0.83%.

This is 0.13% higher than the median reduction in HbA1c of 0.70% achieved by the doctors. On the Hartford cohort, the recommendation of this pipeline is predicted to lead to a median decrease in HbA1c of 0.55%, which is the same as the median reduction in HbA1c achieved by the doctors.

## 4. Conclusion

Using these novel interpretable pipelines, we have created a decision flow for the prescription of treatments of type 2 diabetes. Compared to the current clinical guidelines which are not precise, are unstructured, and are created using consensus among practitioners, our pipelines are precise (different branches for different physiologies of a patients), structured (clear decision-making flow), and created using data. An additional observation is that simply balancing the dataset is usually not enough for interpretable and consistent (with biology) machine learning, at least in this context. It is important to carefully identify and remove outliers from the majority class as well as the minority class.

There are also a few limitations of this study. Out of the patients who experienced an increase in HbA1c after a treatment regimen (including no medication), there was no way to differentiate between those patients who did so because of negligence or non-adherence, and those who did so because the treatment regimen was in-effective for them. Even though, we try to circumvent this hurdle by filtering these patients out, it is important to consider the patients who experienced an increase in HbA1c levels even after correctly following their doctor's prescription. Additionally, we treat each patient visit as an independent event. In future studies, embeddings from the time series of HbA1c and BMI could also be added into the feature space to make better predictions.

But despite these limitations, these pipelines are interpretable and shown to be effective on external data too. There are two additional benefits to the approach suggested. Firstly, it is easily extendable to include other features (like family history of type 2 diabetes): each of the models can be re-trained at a low cost as they are low depth decision trees. Secondly, these pipelines are easily implementable as a set of if-then conditions. Therefore, this system can not only be deployed on EMRs in hospitals and clinics, but also in smartphones and in future medical wearable devices.

To validate this study further, it could be used in pilot studies at hospitals, under the expert supervision of doctors. Newer (but more complex and less interpretable) techniques like holistic AI to incorporate doctor's notes and time series data of BMI and HbA1c, along with tabular data could be used.

We hope that these methods proposed above, provide a useful starting point to create AI-backed interpretable policies for precise type 2 diabetes treatment. This will help us ensure that patients are given the most effective medication according to their unique needs, which will ultimately lead to a healthier society.

## 5. Acknowledgments

Funding: No funding was received.

Duality of Interest: There is no conflict of interest related to this work.

Author Contributions: D.K.A. contributed to defining the problem, developing, suggesting, and implementing methods, and writing, reviewing and editing the paper. D.J.B. contributed to defining the problem, developing and suggesting methods, and reviewing and editing the paper.